\documentclass[10pt,twocolumn,letterpaper]{article}

\usepackage{iccv}
\usepackage{times}
\usepackage{epsfig}
\usepackage{graphicx}
\usepackage{amsmath}
\usepackage{amssymb}

\usepackage{bm}

\usepackage[usenames,dvipsnames]{color}

\newcommand{\mtxt}[1]{\textrm{\normalfont\emph{#1}}}
\newcommand{\isyn}{\ensuremath{I_{\mtxt{syn}}}}
\newcommand{\iobs}{\ensuremath{I_{\mtxt{obs}}}}

\newcommand{\myvec}[1]{\bm{\mathrm{#1}}}
\newcommand{\mymat}[1]{\bm{\mathrm{#1}}}

\usepackage[inline]{enumitem} 

\usepackage{tabularx} 
\usepackage{multirow}
\newcolumntype{Y}{>{\centering\arraybackslash}X}

\usepackage{microtype}

\newcommand{\sysname}{GazeDirector}

\usepackage[numbers]{natbib}

\usepackage{stackengine}
\newcommand{\inlinelabel}[2]{
   \setlength{\fboxsep}{1pt}
    \stackinset{l}{-1pt}{b}{-1pt}{\makebox[0pt][l]{\colorbox{white}{\small(#1)}}}{#2}
}

\usepackage[pagebackref=true,breaklinks=true,letterpaper=true,colorlinks,bookmarks=false]{hyperref}

\iccvfinalcopy

\def\iccvPaperID{X}

\ificcvfinal\fi
\begin{document}

\title{GazeDirector: Fully Articulated Eye Gaze Redirection in Video}

\author{Erroll Wood\textsuperscript{1,4} \hspace{0.45em} Tadas Baltru\v{s}aitis\textsuperscript{2} \hspace{0.45em} Louis-Philippe Morency\textsuperscript{2} \hspace{0.45em} Peter Robinson\textsuperscript{1} \hspace{0.45em} Andreas Bulling\textsuperscript{3} \\
\textsuperscript{1}University of Cambridge, UK \quad \textsuperscript{2}Carnegie Mellon University, USA \\ \textsuperscript{3}Max Planck Institute for Informatics, Germany \quad \textsuperscript{4}Microsoft
}

\maketitle


\begin{abstract}

We present \sysname{}, a new approach for eye gaze redirection that uses model-fitting.
Our method first tracks the eyes by fitting a multi-part eye region model to video frames using analysis-by-synthesis, thereby recovering eye region shape, texture, pose, and gaze simultaneously.
It then redirects gaze by 1) warping the eyelids from the original image using a model-derived flow field, and 2) rendering and compositing synthesized 3D eyeballs onto the output image in a photorealistic manner.
\sysname{} allows us to change where people are looking without person-specific training data, and with full articulation, i.e. we can precisely specify new gaze directions in 3D.
Quantitatively, we evaluate both model-fitting and gaze synthesis, with experiments for gaze estimation and redirection on the Columbia gaze dataset.
Qualitatively, we compare \sysname{} against recent work on gaze redirection, showing better results especially for large redirection angles.
Finally, we demonstrate gaze redirection on YouTube videos by introducing new 3D gaze targets and by manipulating visual behavior.\footnote{\url{https://www.youtube.com/watch?v=-tDaZk9V1Nw}}

\end{abstract}


\section{Introduction}

\begin{figure}
    \includegraphics[width=\linewidth]{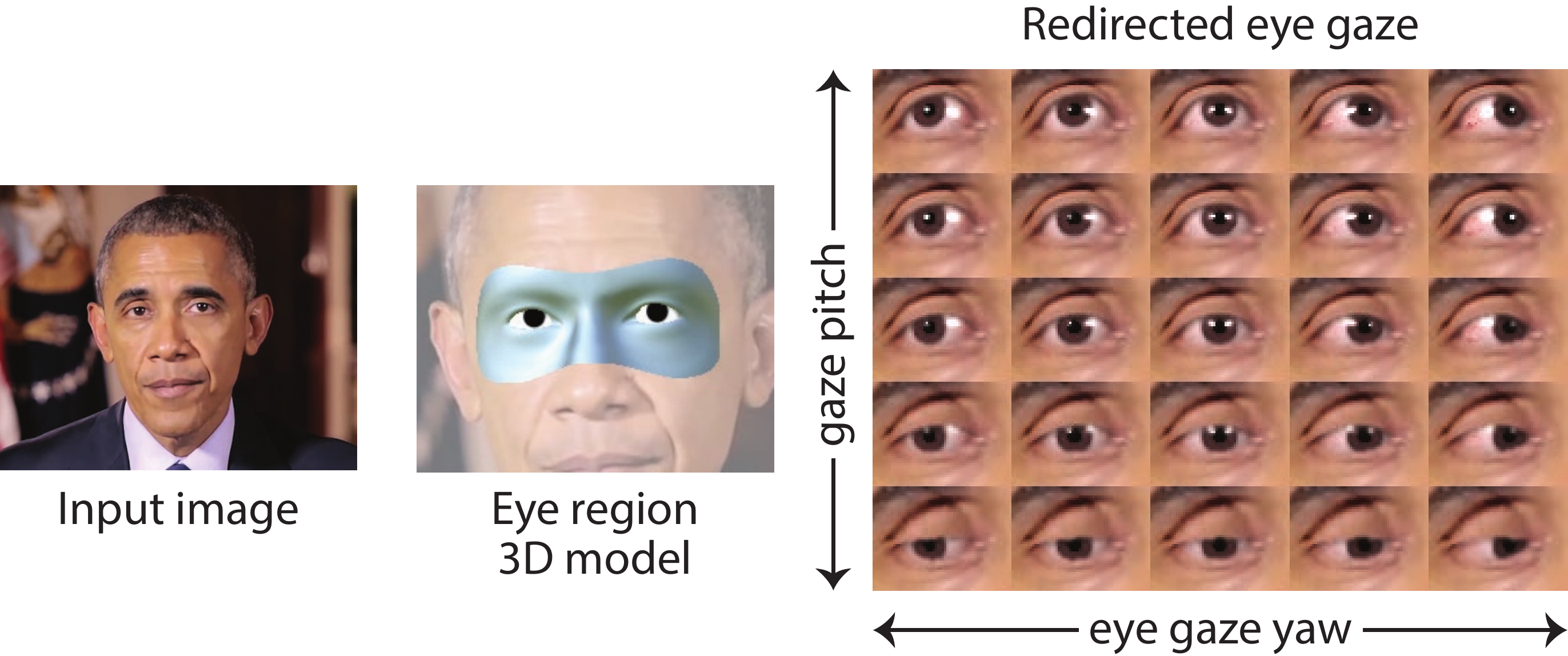}
    \caption{GazeDirector is a new 3D model based approach for gaze redirection. We first recover the shape and appearance of the eyes by fitting a 3D eye region model. We then redirect gaze by warping the eyelids and rendering new redirected eyeballs. Examples of redirected gaze can be seen on the right.}
    \label{fig:teaser}
\end{figure}

Gaze redirection is an upcoming research topic in computer vision where the goal is to alter an image to change where someone appears to be looking (see~\autoref{fig:teaser})~\cite{ganin2016deepwarp,thies2016facevr}.
This is an important generalization of the classic gaze correction problem~\cite{criminisi2003gaze,zitnick1999manipulation}, in which someone's gaze is adjusted along a single direction to simulate eye contact.
With gaze redirection, gaze can be adjusted in any direction.

\begin{figure}
    \inlinelabel{a}{\includegraphics[width=0.49\linewidth]{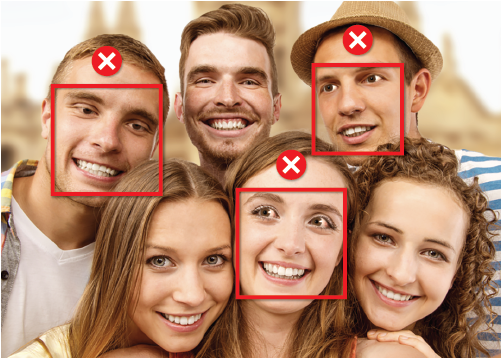}} \hspace{-2em} \hfill
    \inlinelabel{b}{\includegraphics[width=0.49\linewidth]{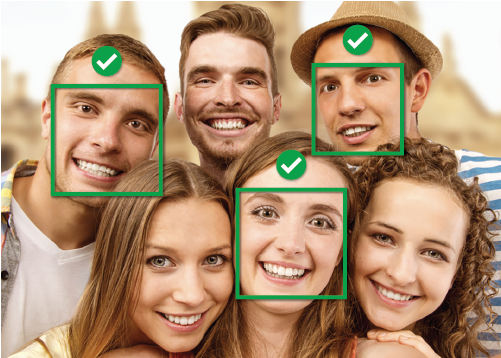}} \par \smallskip
    \inlinelabel{c}{\includegraphics[width=0.49\linewidth]{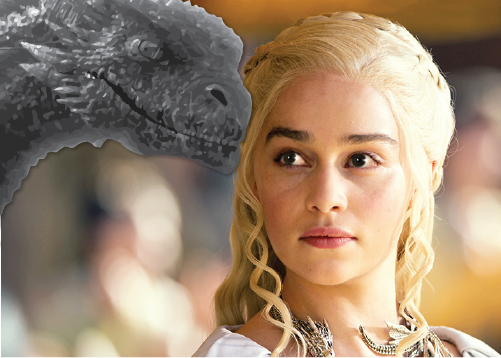}} \hspace{-2em} \hfill
    \inlinelabel{d}{\includegraphics[width=0.49\linewidth]{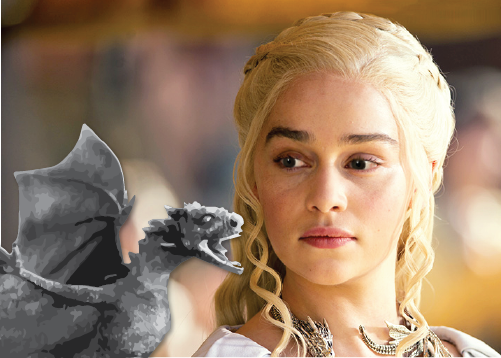}}
    \caption{GazeDirector enables new applications that were previously impossible. (a) Taking group pictures with everyone looking at the camera can be tricky. (b) A gaze correcting camera can ensure this is always the case. (c,d) A challenge for actors is knowing where to look before visual effects are added to a scene. This can be modified in post-production, so if a CGI character is changed, the actor's gaze can be adjusted accordingly. The highlighted faces in (a,b) and the face in (c,d) have been modified by \sysname{}.}
    \label{fig:applications}
\end{figure}

The ability to freely change where someone is looking paves the way for a variety of compelling new applications (see~\autoref{fig:applications}).
For example, taking a group picture with everyone is looking at the camera at the same time can be difficult~\cite{Smith2013}.
Imagine a gaze-correcting camera that could always enforce eye contact, no matter where people are actually looking.
Also, one challenge for actors nowadays is performing alone before other computer-generated characters are composited in.
Where are they supposed to look?
With gaze redirection their apparent point-of-regard could be controlled in post-production, ensuring they look at virtual characters.
Gaze direction is also an important social signal~\cite{emery2000eyes} -- the ability to redirect gaze or even impose specific visual behaviours on video content in real-time could serve as a useful experimental tool, e.g.\ to study gaze following or joint attention in autism research~\cite{jones2004joint}.

A reliable and robust gaze redirection algorithm should work with previously unseen people and handle desired gaze directions which differ significantly from the original gaze.
\citet{thies2016facevr} recently proposed an approach which requires per-user calibration -- a tedious process that is unsuitable for many scenarios.
More relevant to our goal of user-independent gaze redirection is DeepWarp \cite{ganin2016deepwarp}, an approach that uses a deep neural network to directly predict an image-warping flow field between two eye images with a known gaze ``correction'' angular offset between them.
This flow field is applied to the original image to redirect gaze.
In this way, DeepWarp can only redirect gaze by shifting it by an angular offset; it cannot specify new gaze directions explicitly.
Furthermore,
this approach is prone to producing unsightly artefacts when redirecting gaze over large angles.
This problem is fundamental in any purely warping-based approach since it is impossible to warp parts of the eye that were occluded in the original image.

In this work we present \emph{\sysname{}}, a new approach for person-independent gaze redirection.
The main idea of our approach is to model the eye region in 3D instead of trying to predict a flow field directly from an input image~\cite{ganin2016deepwarp}.
Since we recover the shape and pose of the eyes in 3D, our approach can redirect gaze with \emph{full articulation}: \sysname{} can precisely specify new desired gaze targets or directions in 3D instead of using gaze angle correction offsets~\cite{ganin2016deepwarp}.
To model the eye in 3D, we extend a recently proposed method~\cite{wood20163d} to fit a 3D morphable model of the eye region to both eyes in an input image using analysis-by-synthesis.
Once we have recovered the 3D shape, pose, and appearance of the eyes we redirect gaze in two steps.
First, we compute a dense model-derived flow field corresponding to eyelid motion between the original and desired gaze directions.
This dense flow field is efficiently extrapolated from sparse per-vertex flow values using GPU rasterization. 
We apply this flow field to the input image to warp the eyelids.
Second, we render and composite our redirected eyeball models onto the output image in a photorealistic manner.

\medskip
\textbf{Contributions} \quad
1) Our primary contribution is \sysname{} -- a new method that demonstrates how eye-region model fitting using analysis-by-synthesis enables superior gaze redirection compared to previous approaches~(\S\ref{sec:overview}).
In addition, we present the following secondary contributions:
2) A practical approach for rapid synthesis of dense model-derived optical flow fields using GPU rasterization~(\S\ref{subsec:warping-the-eyelids}).
3) Improvements over the state-of-the-art in gaze estimation using our dataset-independent model fitting approach~(\S\ref{sec:columbia}).

\section{Related Work}

\textbf{Eye gaze manipulation} \quad
The lack of eye contact during video-conferencing is a well-known problem.
In computer vision, there are three main approaches to tackle it:
1) novel-view synthesis, 
2) eye-replacement, and 
3) eye-warping.

Novel-view synthesis methods re-render the subject's face so they appear to be looking at the camera.
The first step is recovering a dense depthmap of the face -- this can been done with stereo vision \cite{criminisi2003gaze,yang2002eye}, RGB-D (color with depth) cameras \cite{kuster2012gaze}, and monocular RGB cameras \cite{giger2014gaze}.
This facial depthmap is then rotated and re-rendered from a new viewpoint along a frontal gaze path.
However, as these methods distort the face as a whole, they are not suitable for more general forms of of gaze manipulation.

Eye-replacement methods replace eyes in the original image with new eye images representing different gaze.
The most realistic approaches collect a set of person-specific images of eyes looking at a camera, and composite them into the original face \cite{qin2015eye,shu2016eyeopener,wolf2010eye}.
These methods require person-specific eye images to pick from, and encounter issues when compositing eyes across different head poses or illumination conditions.
Other eye-replacement approaches synthesize new eyeballs with graphics \cite{gemmell2000gaze,weiner2003virtual}.
However, these methods do not move the eyelids -- an important cue for vertical gaze, and only use rudimentary 2D graphics techniques that ignore iris color, head pose, and scene illumination.
Our method instead synthesizes new eyeballs taking eyelid motion, iris color, and illumination into account.

Warping-based methods can redirect gaze without requiring person-specific training data.
These methods learn to generate a flow field from one eye image to another using training pairs of eye images with known gaze offsets between them.
This flow field is used to warp pixels in the original image, thus modifying gaze \cite{ganin2016deepwarp,kononenko2015learning}.
However, purely warping-based methods suffer three major limitations:
First, they can only offset the original unknown gaze direction, so cannot specify a new gaze direction explicitly.
Second, the range of possible redirection is limited by the gaze directions in the training set.
Third, warping artefacts appear for large redirection angles as parts of the eye that were originally occluded cannot be synthesized correctly.
Using 3D models, \sysname{} can explicitly specify new gaze directions in 3D, without training data, and without introducing artefacts.

Like us, \citet{banf2009example} used morphable models to redirect gaze.
They fit a single-part face model to an image, and redirect gaze by deforming the eyelids using an example-based approach, and sliding the iris across the model surface using texture-coordinate interpolation.
Since they use a mesh where the face and eyes are joined, their method only works when people look straight ahead. 
\sysname{} instead models the face and eyeballs as separate parts, letting it work for non-frontal input gaze and allowing the eyeball to rotate separately from the eyelids, as it does in real life.

\begin{figure*}
    \includegraphics[width=\linewidth]{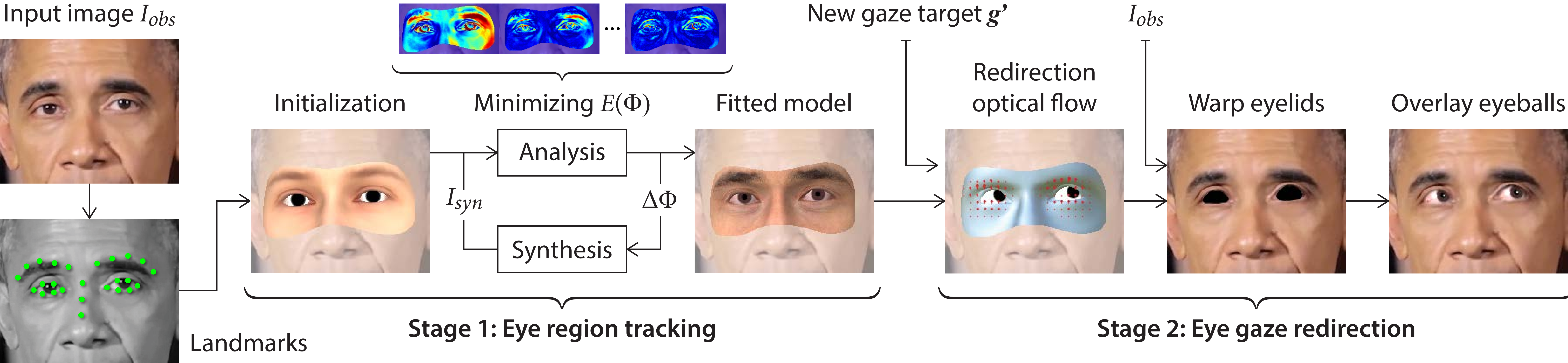}
    \caption{Given observed image \iobs{}, we first initialize our model using 25 facial landmarks from a face tracker \cite{Baltrusaitis2016}. We then find optimal model parameters $\bm\Phi^*$ using analysis-by-synthesis, minimizing a reconstruction energy $E(\bm\Phi)$. We then modify $\bm\Phi^*$ with the desired gaze and eyelid behaviour, resulting in a new \isyn{} which we blend onto \iobs{}, giving a redirected gaze image.}
    \label{fig:overview}
\end{figure*}

\medskip
\textbf{Facial performance capture} \quad

Since \sysname{} recovers the shape, texture, pose, and gaze of the facial eye region, it is also related to work on monocular facial perfomance capture -- a well established research topic \cite{klehm2015recent}.
The goal is to recover dynamic facial geometry and appearance using commodity cameras alone.

Monocular facial performance capture is a highly under-constrained problem, so a parametric face model \cite{blanz1999morphable} is often used as a prior to help recover shape and albedo.
Such models can then be fit to either RGB-D data \cite{thies2015real,weise2011realtime} or RGB data \cite{cao2014displaced,suwajanakorn2015makes,thies2016face}.
However, these approaches generally avoid the eyes, cutting them out of the mesh \cite{cao2015real,thies2016face}.
This is because the parametric face model they use only represents the surface of the skin, and has reduced fidelity around the eye due to poor correspondences in the source head scan data.
For \sysname{}, we extended a previous model that was built using high quality scans \cite{wood20163d}, with care taken to maintain correspondences around the eyelids and eye corners.
Critically, this model treats the eyeballs as separate parts that move independently from the face.

Some previous work tracked the eyes as a part of the face.
\citet{GZCVVPT16} include eyeball geometry in a ``detail'' layer of their facial mesh.
Though this can lead to acceptable re-rendering, it does not allow gaze redirection as the eyeballs and face are joined in a single mesh.
\citet{suwajanakorn2015makes} model eyeball movement by interpolating between facial textures.
This does not allow smooth arbitrary eyeball motion, and requires a large training set of person-specific images with eye movement.
Recent work has combined a facial skin surface capture system with a separate gaze tracker \cite{thies2016face,wang2016realtime,cao2016real}.
Our approach instead captures the facial eye region and eyeball simultaneously. 
This lets us reliably recover eyeball shape and texture parameters -- important for realistic gaze redirection.

There have been recent breakthroughs in capturing the eyeballs and eyelids in extreme detail using special equipment~\cite{berard2014high,berard2016lightweight,Bermano:2015:DSR:2809654.2766924}.
Our work does not come close to this level of detail.
Instead, we focus on capturing the eye for gaze redirection in commodity monocular images and video.


\section{Overview}
\label{sec:overview}

As shown in \autoref{fig:overview},
our approach consists of two main stages: eye region tracking and eye gaze redirection.

\textbf{Tracking} \quad
Given a monocular RGB image frame, we first capture the eyes by fitting our eye region model.
This model consists of two parts: a generative facial part and an articulated eyeball part.
It is defined by a set of parameters $\bm\Phi$ that describe shape, texture, pose, and scene illumination.
We fit our model to the image using analysis-by-synthesis, searching for optimal parameters $\bm\Phi^*$ by minimizing a photometric reconstruction energy.

\textbf{Redirection} \quad
We redirect gaze in two steps:
1) We warp the eyelids in the original image using a flow field derived from our 3D model.
We efficiently calculate this flow field by re-posing our eye region model to change gaze, and rendering the image-space flow between tracked and re-posed eye regions.
2) We then render the redirected eyeballs and composite them back into the image.
We blur the boundary between the skin and eyeball to soften the transition.they ``fit in'' better.


\section{Eye region tracking}
\label{sec:tracking}

For our gaze redirection to look plausible, we must first recover the original shape and texture of the eye region.
Given an image frame $\iobs{}$, 
we therefore wish to recover a set of optimal parameters $\bm\Phi^*$ that best explains it in terms of our eye region model.
We search for $\bm\Phi^*$ using analysis-by-synthesis: iteratively rendering a synthetic eye region image $\isyn{}$, comparing it to $\iobs{}$ using our reconstruction energy $E$ (defined in~\autoref{eqn:energy_parts}), and updating $\bm\Phi$ accordingly.

\subsection{Eye region model}
\label{subsec:synthesis}

At the heart of our method lies a multi-part eye region model based on that by \citet{wood20163d}.
For \sysname{}, we extended it to model two eyes rather than one, simplified the iris color model to improve robustness, and added aesthetic improvements (subdivision surfaces, ambient occlusion) to improve realism.
Our model contains four main parts: the left and right facial eye regions, and the left and right eyeballs. It is parameterized by $\bm\Phi$:
\begin{equation}
	\bm\Phi = \left\{\bm\beta, \bm\tau, \bm\theta, \bm\iota \right\},
\end{equation}
where $\bm\beta$ are the set of shape parameters, $\bm\tau$ the texture parameters, $\bm\theta$ the pose parameters, and $\bm\iota$ the illumination parameters.
We now describe each parameter below.

\begin{figure}[t]
	\centering
	\begin{tabularx}{1.0\linewidth}{@{}Y@{}Y@{}}
		Shape model $\mathcal{M}_\mtxt{geo}$ & Texture model $\mathcal{M}_\mtxt{tex}$
	\end{tabularx} \\ \smallskip
    \includegraphics[width=1.0\linewidth]{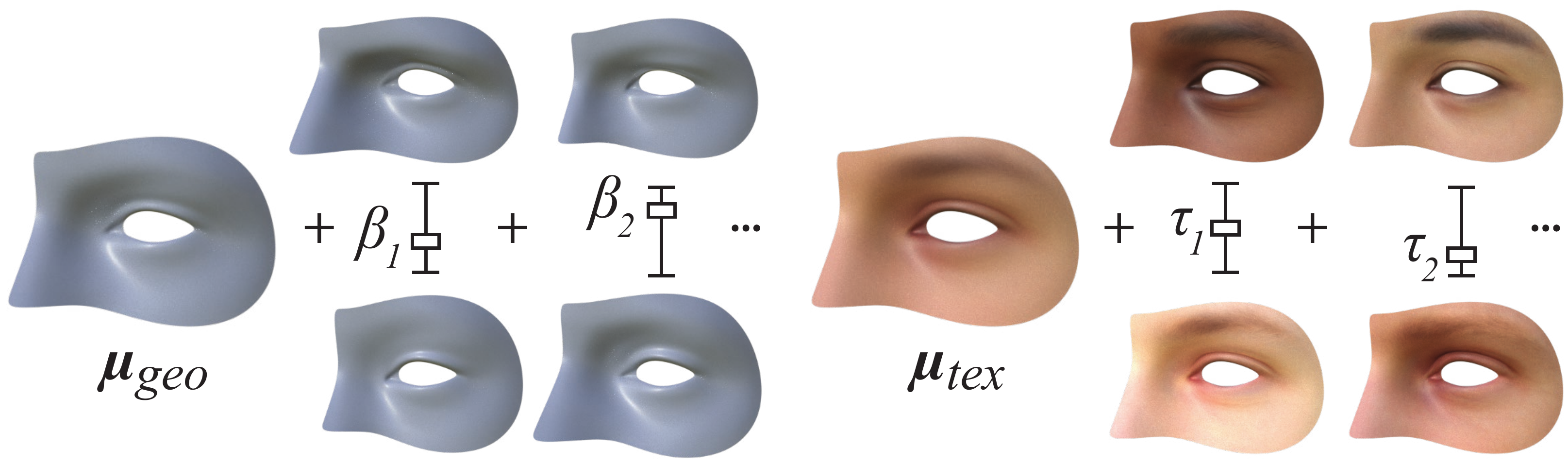}
    \caption{The average facial shape $\bm{\mu}_\mtxt{geo}$ and texture $\bm{\mu}_\mtxt{tex}$, along with the top modes of variation. The first mode of shape variation moves between hooded and protruding eyes, and the first mode of texture variation moves between dark and light skin.}
    \label{fig:pca-examples}
\end{figure}

\medskip
\textbf{Shape $\bm\beta$} \quad
The geometric shape of each eye region is described by a linear Principal Component Analysis (PCA) model $\mathcal{M}_\mtxt{geo} \!\in\!\mathbb{R}^{3n}$ in the style of previous work \cite{blanz1999morphable}.
This comprises $n\!=\!229$ vertices and was built from a collection of 22 high resolution scans acquired online \cite{wood16_etra}.
We assume faces are symmetrical, so the shapes of both eye regions are controlled with a single set of coefficients $\bm\beta_{\mtxt{face}}\!\in\!\mathbb{R}^{16}$,
\begin{equation}
	\mathcal{M}_\mtxt{geo}(\bm\beta_{\mtxt{face}}) = \bm\mu_\mtxt{geo} + \mymat{U} \textrm{diag}(\bm\sigma_\mtxt{geo}) \bm\beta_{\mtxt{face}}
	\label{eq:shape}
\end{equation}
where $\bm\mu_\mtxt{geo}$ is the average face shape, $\mymat{U}$ the modes of shape variation, and $\bm\sigma_\mtxt{geo}$ the standard deviations of these modes (see \autoref{fig:pca-examples}).
For simplicity, each $\beta_i\!\in\!\bm\beta_{\mtxt{face}}$ is scaled so that $\beta_i\!=\!1$ represents one standard deviation's worth of variation in that dimension.
For the eyeball we use a standard two-sphere model based off physiological averages \cite{ruhland2014look}.
We also include a parameter $\beta_{\mtxt{iris}}$ that controls iris size by scaling vertices on the iris boundary about the pupil.

\begin{figure}[t]
    \includegraphics[width=1.0\linewidth]{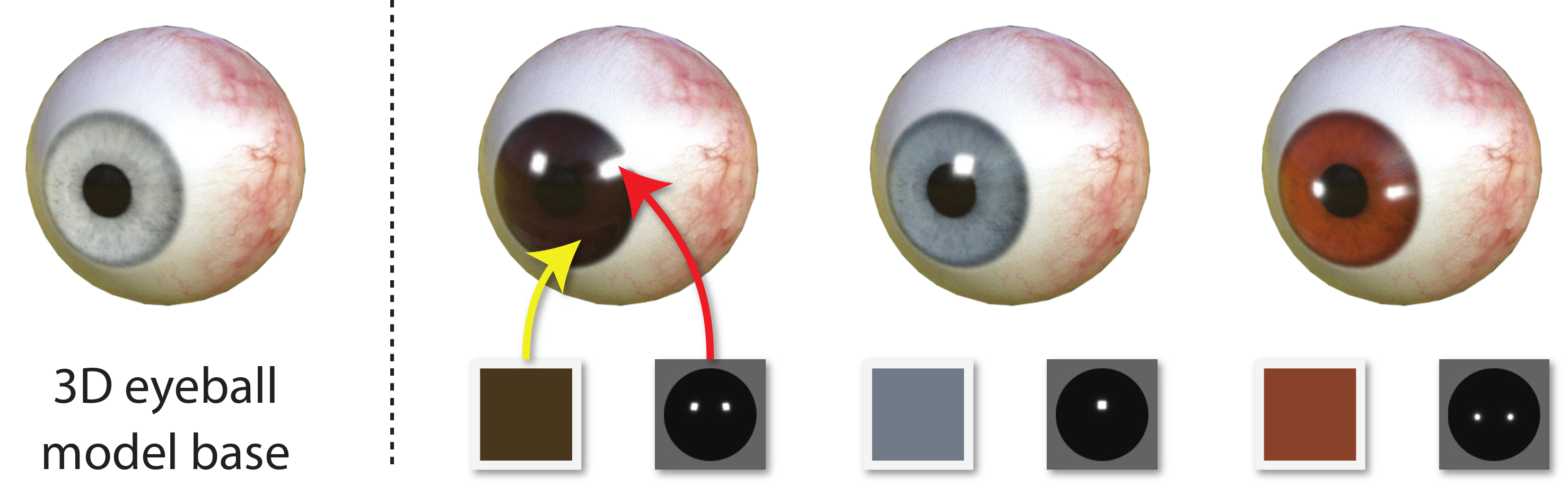}
    \caption{Our eyeball model captures iris color variation with an RGB color $\bm\tau_\mtxt{iris}$ (yellow arrow). Environmental reflections are added with spherical environment maps (red arrow).}
    \label{fig:eyeball}
\end{figure}

\medskip
\textbf{Texture $\bm\tau$} \quad
We use a linear PCA texture model $\mathcal{M}_\mtxt{tex} \!\in\!\mathbb{R}^{3m}$ of the facial eye region, built from the same set of scans.
Rather than model the color of each vertex \cite{blanz1999morphable},
$\mathcal{M}_\mtxt{tex}$ generates RGB texture maps sized $m\!=\!512\!\times\!512$px that we apply to both eye regions.
This linear texture model is controlled with texture coefficients $\bm\tau_{\mtxt{face}}\!\in\!\mathbb{R}^{8}$,
\begin{equation}
	\mathcal{M}_\mtxt{tex}(\bm\tau_{\mtxt{face}}) = \bm\mu_\mtxt{tex} + \mymat{V} \textrm{diag}(\bm\sigma_\mtxt{tex}) \bm\tau_{\mtxt{face}}
	\label{eq:texture}
\end{equation}
where $\bm\mu_\mtxt{tex}$ is the average face texture, $\mymat{V}$ the modes of texture variation, and $\bm\sigma_\mtxt{tex}$ the respective standard deviations.
Each coefficient is scaled in a similar way to $\mathcal{M}_\mtxt{geo}$, so it represents one standard deviation in its dimension.
As shown in \autoref{fig:eyeball}, we vary the iris by multiplying the iris region of the base eyeball texture with an RGB color $\bm\tau_\mtxt{iris}$.
Since the ``white of the eye'' is rarely purely white, we also tint it with another color $\bm\tau_\mtxt{tint}$

\medskip
\textbf{Pose $\bm\theta$} \quad
Our pose parameters describe both global and local pose.
Globally, the eye regions are positioned with rotation $\theta_{\bm{R}}$ and translation $\theta_{\bm{T}}$.
The interocular distance is controlled via $\theta_\mtxt{iod}$
The eyeball positions are fixed in relation to the eye regions.
Our local pose parameters allow the eyeballs to rotate independently from the face, controlling gaze.
The general gaze direction is given by pitch and yaw angles $\theta_p$ and $\theta_y$, and
vergence is controlled with $\theta_v$.
When the eyeball looks up or down, the eyelids follow it.
We use procedural animation to pose the eyelids in the facial mesh by rotational ammount $\theta_\mtxt{lid}$ \cite{wood16_etra}.

\begin{figure}[t]
	\fboxsep0pt
    \fbox{\includegraphics[width=0.235\linewidth]{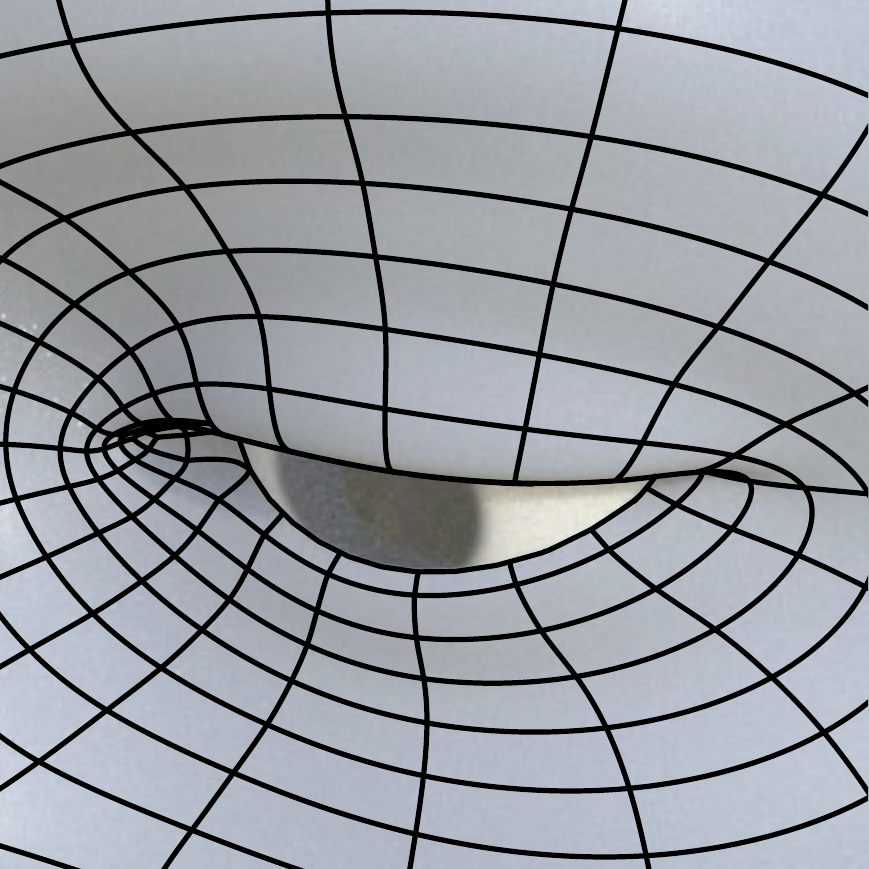}}\hfill
    \fbox{\includegraphics[width=0.235\linewidth]{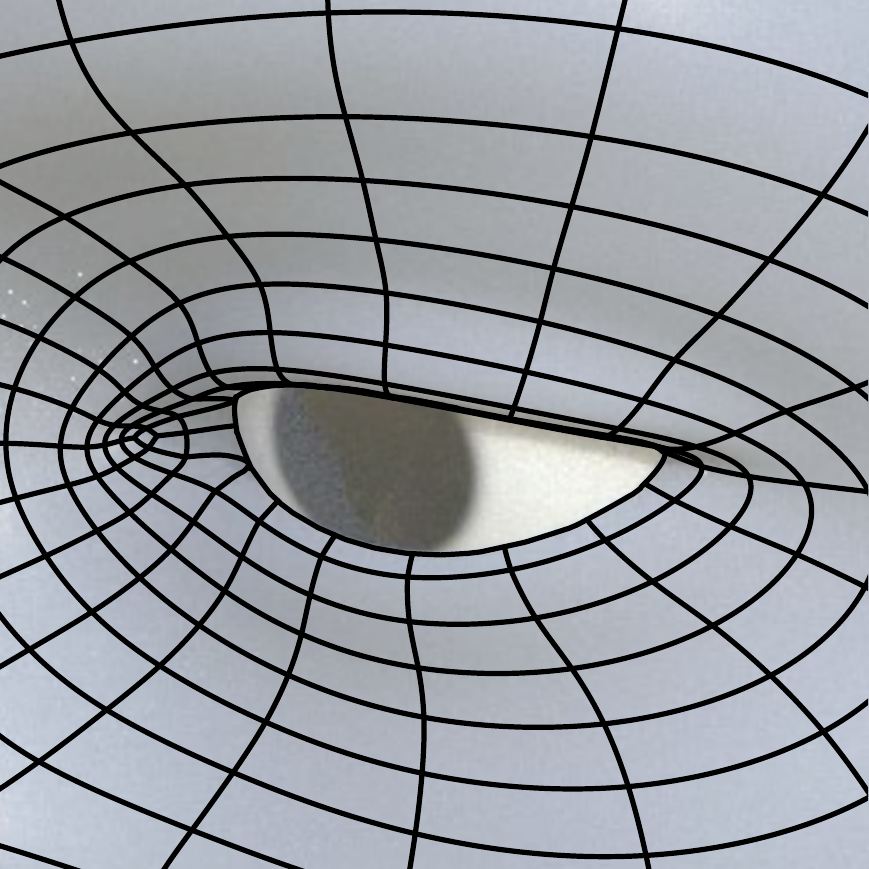}}\hfill
    \fbox{\includegraphics[width=0.235\linewidth]{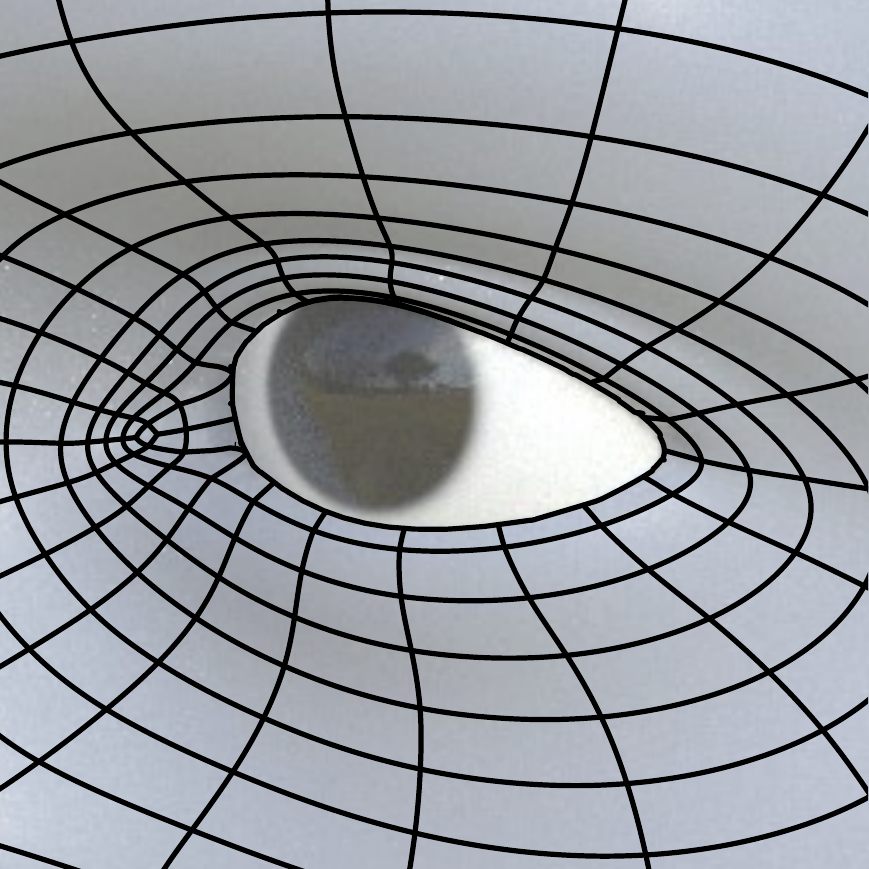}}\hfill
    \fbox{\includegraphics[width=0.235\linewidth]{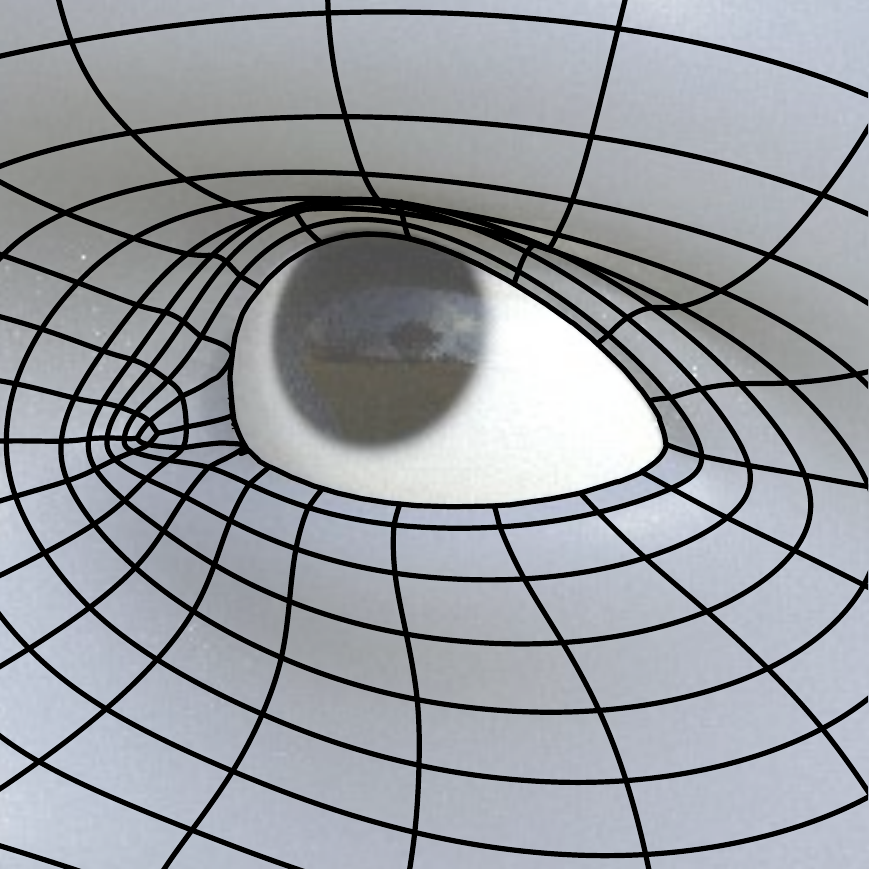}}
    \caption{Our eyelid posed using procedural animation for eyelid gaze pitch angles $\theta_\mtxt{lid}$ between $-20^\circ$ and $+20^\circ$.}
    \label{fig:lidpose}
\end{figure}

\medskip
\textbf{Illumination $\bm\iota$} \quad
We assume a simple illumination model of ambient light coupled with a single directional light.
The ambient light has intensity ${\bm\iota}_{\mtxt{amb}}\!\in\!\mathbb{R}^{3}$, and the directional light has intensity ${\bm\iota}_{\mtxt{dir}}\!\in\!\mathbb{R}^{3}$ and direction defined by rotation ${\bm\iota}_{R}\!\in\!\mathbb{R}^{2}$ (pitch and yaw angles).
We assume all surfaces are Lambertian.
Though $\bm\iota$ cannot describe complex scene illumination, we found it was sufficient in many cases considering the small facial region that we consider.

\medskip
In total we have $17+14+11+9=51$ parameters of $\bm\Phi$ to optimize over.

\begin{figure*}
    \includegraphics[width=\linewidth]{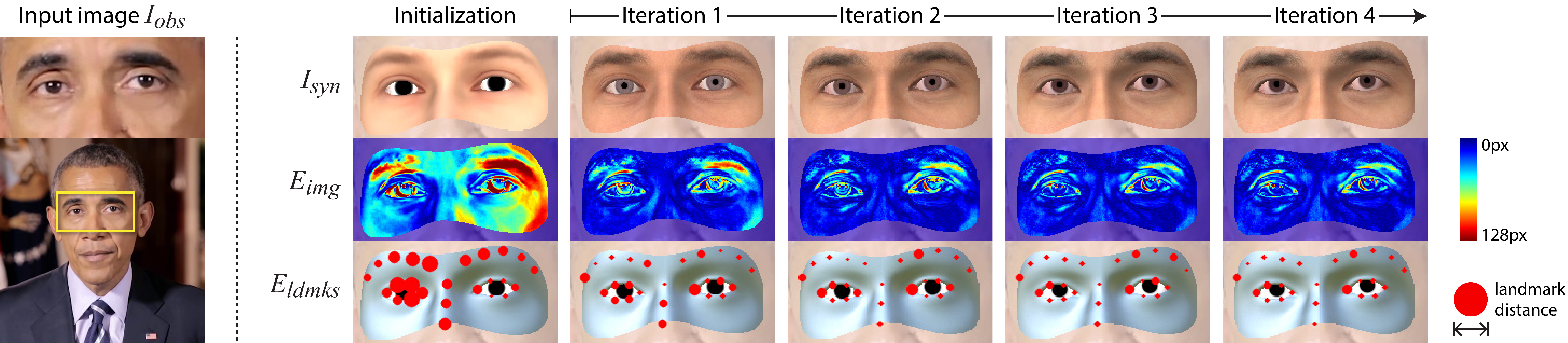}
    \caption{We fit our 3D eye region model to an image my minimizing a reconstruction energy $E(\bm\Phi)$. Our two main energy terms are a dense photometric error term $E_{\mtxt{img}}$ and a sparse landmark similarity term $E_{\mtxt{ldmks}}$. This figure shows the energies decreasing over four iterations of the Gauss Newton algorithm.}
    \label{fig:fitting_iters}
\end{figure*}

\medskip
\textbf{Rendering the model} \quad
Once our model has been configured with parameters $\bm\Phi$, we render synthetic images $\isyn{}(\bm\Phi)$ using a DirectX-based rasterizer.
We fix our virtual camera location at the world origin, and assume knoweldge (or estimate) of camera intrinsic parameters.

Realistically rendering eyes is a challenge \cite{ruhland2014look}.
We implement three additional effects to improve the realism of our output.
First, as our model is low-resolution, it appears blocky when rendered.
We therefore smooth the skin's surface using a single step of Loop subdivision \cite{loop1987smooth} with precomputed stencils for efficiency.
Second, we use physically correct corneal refraction techniques in the eyeball shader to better model its layered transparent structure \cite{ActiBlizEyes}.
Third, we approximate ambient occlusion shadowing on the eyeball using a single-pass analytic techniqe:
we project the positions of eyelid vertices into eyeball $uv$ space, fit a 2D cubic polynomial to them, and apply per-pixel ambient occlusion as a function of distance to each eyelid polynomial.

\subsection{Energy formulation}

A good energy function is critical to the success of any analysis-by-synthesis method.
Our proposed energy $E(\bm\Phi)$ is a weighted sum of several terms, each encoding a different requirement of our model fit.
Each term is expressable as a sum-of-squares, allowing us to minimize $E(\bm\Phi)$ using the Gauss-Newton algorithm.
\begin{equation}
	E(\bm\Phi) = \underbrace{E_{\mtxt{img}}(\bm\Phi)\!+\!E_{\mtxt{ldmks}}(\bm\Phi)}_{\textrm{Data terms}} +
	\underbrace{E_{\mtxt{stats}}(\bm\Phi)\!+\!E_{\mtxt{pose}}(\bm\Phi)}_{\textrm{Prior terms}}
	\label{eqn:energy_parts}
\end{equation}
Our data terms (see \autoref{fig:fitting_iters}) guide our model fit using image pixels and facial landmarks, while our priors penalize unlikely facial shape and texture, and eyeball orientations. 
We now describe each term in detail.

\medskip
\textbf{Image similarity $\bm E_{\mtxt{img}}$} \quad
Our primary goal is to minimize the photometric reconstruction error between \isyn{} and \iobs{}.
The data term $E_{\mtxt{img}}$ expresses how well the fitted model explains \iobs{} by densely measuring pixel-wise differences across the images using a robust mean squared error.
We promote image similarity with the term
\begin{equation}
	E_{\mtxt{img}}(\bm\Phi) = \frac{1}{\left|\mathcal{P}\right|} \sum_{p \in \mathcal{P}} \, \rho \big( \left| \isyn{}(p) - \iobs{}(p) \right| \big)^2
\end{equation}
where $\mathcal{P} \! \subset\!\isyn{}$ represents the set of rendered foreground pixels belonging to our 3D model.
The background pixels are ignored.
The robust function $\rho(e) = \min(\sqrt{T}, e)$, for threshold $T$, alleviates the effects of outliers; this is important for recovering iris color in the presence of strong specular highlights on the eye.
\medskip

\textbf{Landmark similarity $\bm E_{\mtxt{ldmks}}$} \quad
The face contains several landmark feature points that can be tracked reliably.
We therefore regularize our dense data term ($E_{\mtxt{img}}$) using a sparse set of landmarks $\mathcal{L}$ provided by a face tracker \cite{Baltrusaitis2016}.
$\mathcal{L}$ consists of 25 points that describe the eyebrows, nose and eyelids.
For each 2D tracked landmark $l \!\in\! \mathcal{L}$, we also compute a corresponding synthesized 2D landmark $l^\prime$ as a linear combination of projected vertices in our shape model.
Facial landmark similarities are incorporated into our energy using
\begin{equation}
	E_{\mtxt{ldmks}}(\bm\Phi) = \lambda_{\mtxt{ldmks}} \cdot \frac{1}{\left|\mathcal{P}\right|} \sum_{i = 0}^{\left|\mathcal{L}\right|} \, \lVert l_i - l^\prime_i \rVert^2
\end{equation}
As landmark distances $\lVert l_i - l^\prime_i \rVert$ are measured in image-space, we normalize the energy by dividing through by foreground area $\left|\mathcal{P}\right|$ to avoid bias from eye region size in the image.
The importance of $E_{\mtxt{ldmks}}$ is controlled with weight $\lambda_{\mtxt{ldmks}}$.

\medskip
\textbf{Statistical prior $\bm E_{\mtxt{stats}}$} \quad 
We penalize unlikely facial shape and texture using a statistical prior~\cite{blanz1999morphable}.
As we assume a normally distributed population, our PCA model parameters should be close to the mean $\bm 0$:
\begin{equation}
	E_{\mtxt{stats}}(\bm\Phi) =
	\lambda_{\mtxt{geo}} \cdot \sum_{i = 0}^{|\mathcal{\bm\beta}|} \beta_{i}^2 + 
	\lambda_{\mtxt{tex}} \cdot \sum_{i = 0}^{|\mathcal{\bm\tau}|} \tau_{i}^2
\end{equation}
Recall that $\beta_i \!\in\! \bm\beta$ and $\tau_i \!\in\! \bm\tau$ are scaled by their respective standard deviations in our model.
This energy helps our fit avoid degenerate facial shapes and texture, and guides its recovery from poor local minima found in previous frames.
The penalties for unlikely shape and texture are weighted separately with $\lambda_{\mtxt{geo}}$ and $\lambda_{\mtxt{tex}}$.

\medskip
\textbf{Pose prior $\bm E_{\mtxt{pose}}$} \quad 
Our final energy penalizes mismatched parameters for eyeball gaze direction and eyelid position.
The eyelids follow eye gaze, so if the eyeball is looking upwards, the eyelids should be rotated upwards, and visa versa.
We enforce eyelid pose consistency with
\begin{equation}
	E_{\mtxt{pose}}(\bm\Phi) = \lambda_{\mtxt{pose}}  \cdot \lVert \theta_{\mtxt{lid}}-\theta_p \rVert^2
\end{equation}
where $\theta_{\mtxt{lid}}$ is the eyelid pitch angle of our model's face parts, and $\theta_p$ is the gaze pitch angle of our eyeball parts.
Its relative importance is controlled by weight $\lambda_{\mtxt{pose}}$.
\subsection{Optimization procedure}

\begin{figure}
    \includegraphics[width=\linewidth]{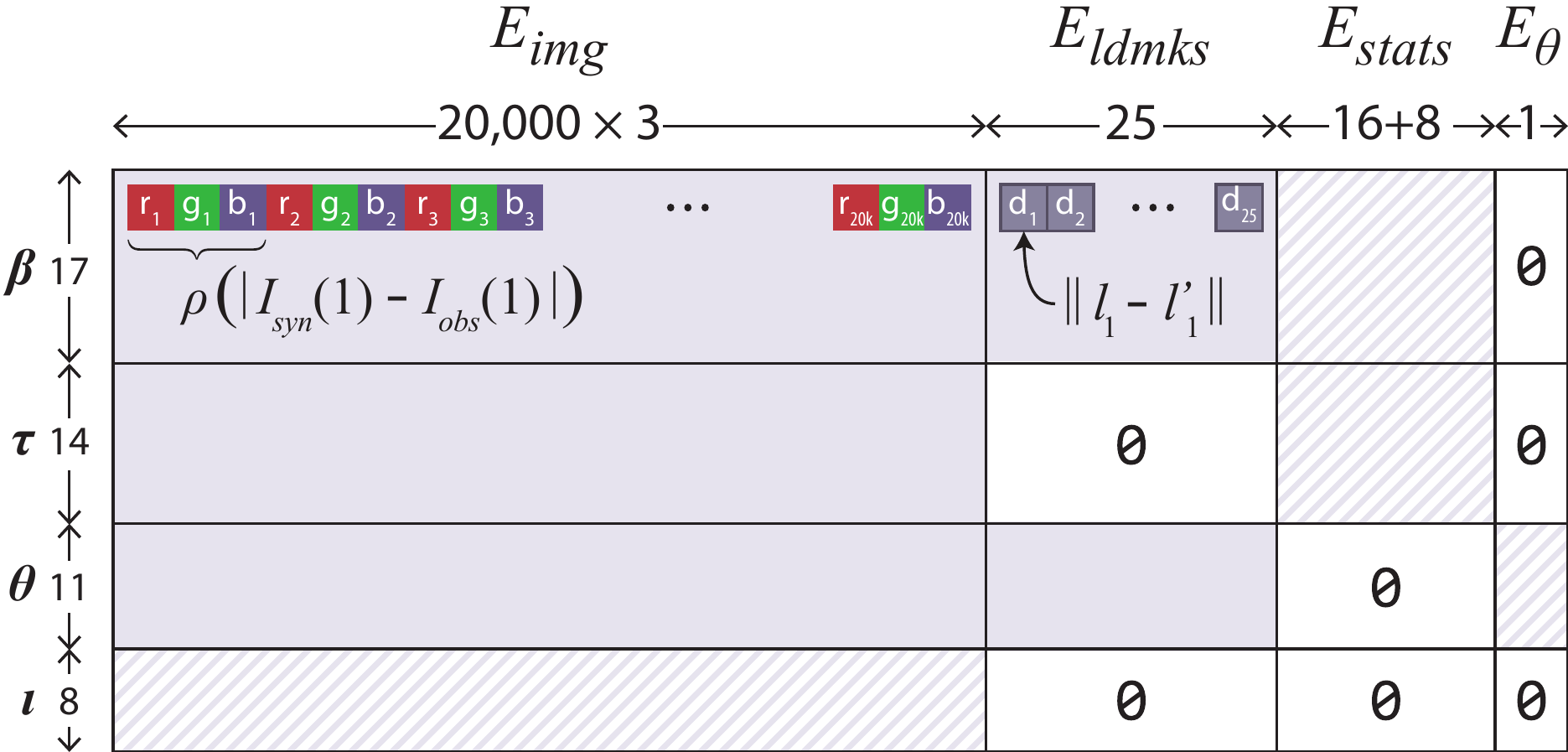} 
    \caption{The non-zero structure of our Jacobian $\mymat{J_r}$ for a $200\!\times\!100$px eye region. $\mymat{J_r}$ is calculated entirely on the GPU. Dashed regions represent sparse blocks.}
    \label{fig:jacobian}
\end{figure}

Minimizing our proposed objective $E(\bm\Phi)$ is a challenging high-dimensional non-convex optimization problem.
We use a GPU-assisted, annealed form of the Gauss-Newton algorithm, where the parameter update for $\bm\Phi$ is as follows:
\begin{equation}
\bm\Phi^{i+1}=\bm\Phi^i  - \eta^i \, (\mymat{J_r}^T \mymat{J_r})^{-1} \cdot \mymat{J_r}^T \myvec{r}
\end{equation}
where $\myvec{r}$ is the vector of energy function residuals, $\mymat{J_r}$ the Jacobian matrix of residuals $\myvec{r}$ evaluated at $\bm\Phi^i$, $\mymat{J_r}^T \mymat{J_r}$ the approxmation to the Hessian matrix, and $\eta$ the annealing rate.
We perform a variable number of Gauss-Newton iterations, terminating early if no more progress is being made.
\autoref{fig:fitting_iters} shows four iterations of our model fit.

To compute the Jacobian we use numerical central derivatives.
This is an expensive operation, requiring two images to be rendered for every parameter.
We keep our system performant by calculating $\mymat{J_r}$ and $\mymat{J_r}^T \mymat{J_r}$ entirely on the GPU, avoiding expensive pipeline stalls from cross-system data transfer.
Additionally, since image rendering is a key operation for our system, we use a tailored DirectX rasterizer that can render \isyn{} over 5000 times per second.
To further lighten the computational load of our numerical derivatives, we mask out a subset of $\bm\Phi$ when tracking in a video, so optimize over a smaller set of parameters frame-to-frame.
As a result, \sysname{} can run at interactive rates.

\medskip
\textbf{Initialization} \quad
The energy landscape of $E(\bm\Phi)$ is riddled with local minima, so we must start from a good initializion.
Our face tracker provides 3D estimates for the facial landmark positions.
We initialize global translation to the mean landmark position and set global rotation parameters using the the \citet{kabsch1976solution} algorithm.
Other parameters are initialized to $\bm 0$ by default, except for interocular distance and iris size, for which we use anthropomorphic averages, and illumination, for which we experimentally chose a basic setup.
When tracking in video, we exploit temporal similarities by initializing $\bm\Phi_{\mtxt{init}}$ with $\bm\Phi^*$ from the previous frame.

\section{Eye gaze redirection}
\label{sec:redirection}

Once we have obtained a set of fitted model parameters $\bm\Phi^*$ for an image $\iobs{}$, our next step is to redirect gaze to point at a new 3D target $\bm{g}^\prime$. 

We first modify $\bm\Phi^*$ to obtain $\bm\Phi^\prime$ that represents the redirected gaze.
We then calculate the optical flow between eye region models with $\bm\Phi^*$ and $\bm\Phi^\prime$, and use this to warp the eyelids in the source image.
Finally, we render the redirected eyeballs and seamlessly composite them into the output image.

\medskip
\textbf{Re-posing our model} \quad
The first step of gaze re-direction is straightforward:
given a new target $\bm{g}^\prime$, we calculate new values for eye gaze pitch $\theta^\prime_p$, yaw $\theta^\prime_y$, and vergence $\theta^\prime_v$ so each eyeball points towards $\bm{g}^\prime$.
Furthermore, we calculate $\theta^\prime_\mtxt{lid}$ to match the new gaze direction.
Altogether, these new gaze parameters are encoded in $\bm\Phi^\prime$.

\begin{figure}
    \includegraphics[width=\linewidth]{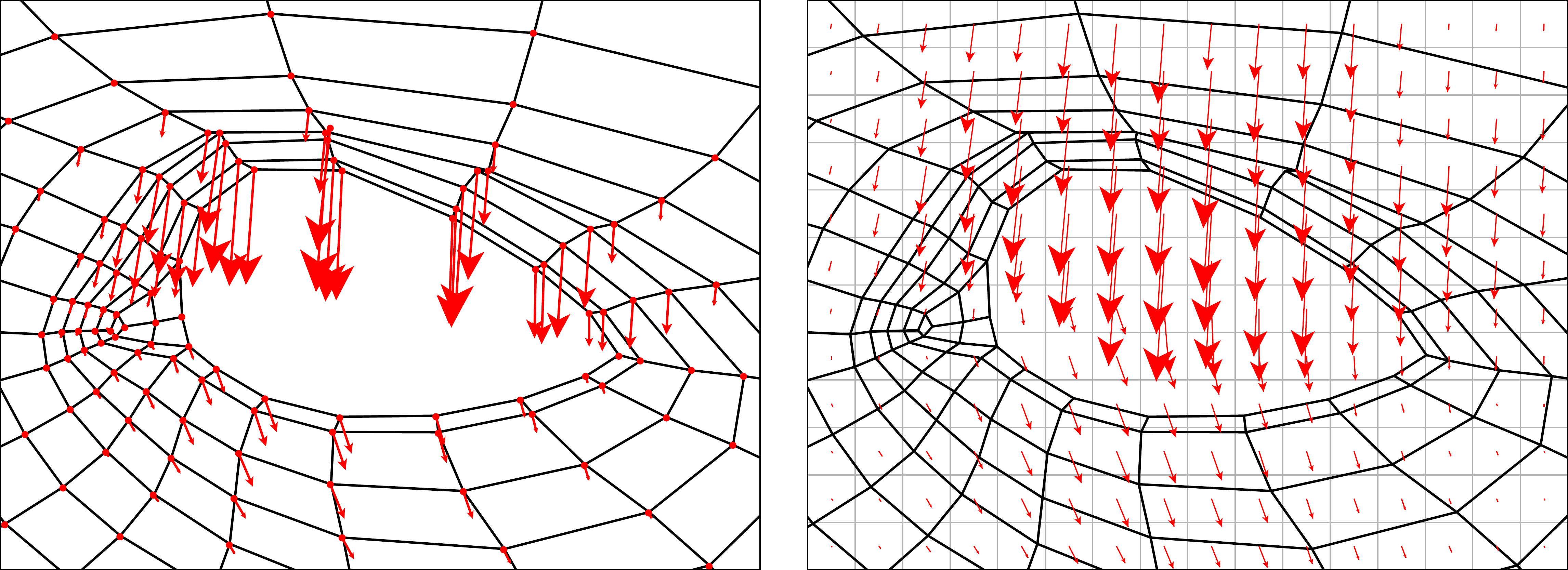}
    \caption{We efficiently convert sparse per-vertex image-space flows (left) to a dense per-pixel flow field (right) using GPU rasterization. We use this dense flow-field to warp the eyelids.}
    \label{fig:flow-field}
\end{figure}

\subsection{Warping the eyelids}
\label{subsec:warping-the-eyelids}

When the eyeball rotates, the eyelids move with it.
To simulate this, we warp the eyelids from the original image using a model-derived optical flow field $\bm{O}$.
To calculate $\bm{O}$, we first calculate the sparse screen-space flow $\bm{o}_i\!\in\!\mathbb{R}^{2}$ for each vertex $\bm{v}_i\!\in\!\mathbb{R}^{3}$ in both facial parts of the eye region:
\begin{equation}
	\bm{o}_i = 
	\Pi \left( \Theta^\prime (\bm{v}_i) \right) - \Pi \left( \Theta^* (\bm{v}_i) \right) \quad i \in [0,458]
\end{equation}
where $\Pi$ is the projection defined by our camera parameters, and $\Theta^{*|\prime}$ are the transforms that combine eyelid motion ($\theta_\mtxt{lid}$) with model-to-world transforms $\theta_{\bm{R}}$ and $\theta_{\bm{T}}$.
It is common for analysis-by-synthesis methods to use GPU rasterization to evaluate an objective function \cite{sharp2015accurate,thies2016face}.
We propose a simple and efficient approach for computing dense flow-fields using the same framework.
To efficiently distribute sparse flow values across image space, we load per-vertex flows $\bm{o}_i$ into our renderer as vertex attributes and let the rasterization stage interpolate between them and handle occlusions between different model parts (see \autoref{fig:flow-field}).
This takes $\sim\!5$ms.
The result is a dense flow field $\bm{O}$ that we use to remap source image pixels to simulate eyelid motion.

\subsection{Compositing redirected eyeballs}

Once the eyelids have been warped, we render the portion of the eyeballs between the eyelids and composite them onto the output image.
Following rasterization, the eyelid edges will be perfectly sharp and unlikely to match the observed image.
We therefore follow the approach adopted by the real-time rendering community \cite{ActiBlizEyes,UnrealEyes}, and blur the seam where the eyeballs meet the eyelids with a small Gaussian.

A shortcoming of our underlying scene model is the lack of specular reflections on the eyeball surface.
Real world eye images often exhibit strong highlights or glints.
We decided not to explicitly model multiple light sources in $\bm\Phi$ because of the additional computational cost with numerical derivatives.
We instead pre-rendered a set of five spherical reflection maps that model common environmental lighting scenarios (see \autoref{fig:eyeball}), and use them to apply specular reflections on the eyeball at runtime.
This choice is made by seeking the reflection map that minimizes image error.
While this cannot model complex environmental reflections, it improves the perceived quality of the eyeball re-rendering.


\section{Evaluations}

In this section we evaluate \sysname{}.
Quantitatively, we evaluate our model fitting stage with a gaze estimation experiment, and our gaze synthesis stage with a gaze redirection experiment.
Qualitatively, we compare our method against recent work and demonstrate gaze redirection and visual behaviour manipulation on YouTube videos.

\begin{figure}
    \includegraphics[width=\linewidth]{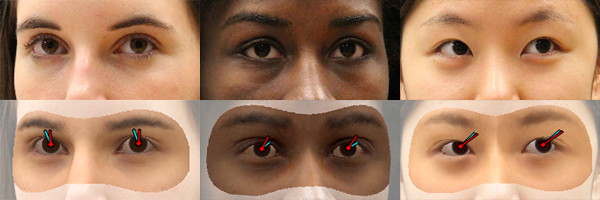} 
    \caption{Eye region model fits on the Columbia gaze dataset \cite{Smith2013} showing true gaze (red) and estimated gaze (cyan).}
    \label{fig:example_fits_columbia}
\end{figure}

\begin{figure}
    \includegraphics[width=\linewidth]{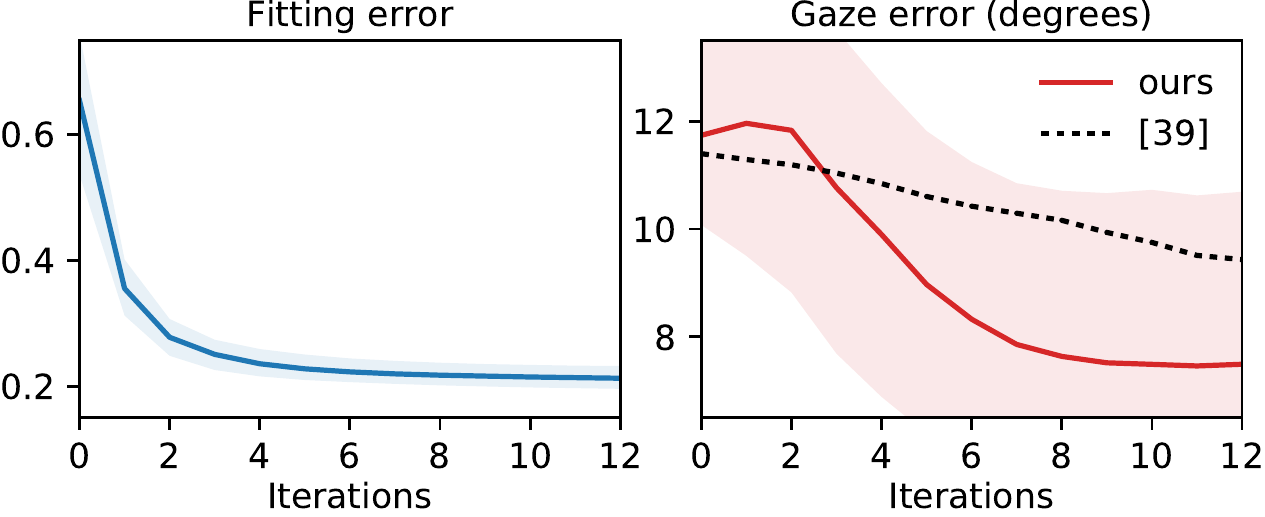} 
    \caption{Fitting error and gaze error for the Columbia dataset \cite{Smith2013} decrease with the number of fitting iterations. Line is median, filled region is interquartile range. Our second-order optimization strategy converges faster than previous first-order methods \cite{wood20163d}.}
    \label{fig:exp-gaze}
\end{figure}

\subsection{Model fitting performance}
\label{sec:columbia}

We performed an experiment to assess our fitting strategy.
We measured two factors:
1) photometric error to determine how well we reconstructed the image, and
2) gaze estimation error to see if we can correctly recover eyeball pose.
We used the Columbia gaze dataset~\cite{Smith2013}, which contains images of 56 people looking at a target grid on the wall. 
The participants were constrained by a head-clamp, and images were taken from five different head orientations. 
In our experiments we used a subset of 34 people (excluding those with eyeglasses) with 20 images per person.

Results of our experiment can be seen in \autoref{fig:exp-gaze}, and example model fits can be seen in \autoref{fig:example_fits_columbia}.
Photometric error and gaze estimation error decrease with the number of model fitting iterations.
This confirms the effectiveness of our fitting strategy.
If we examine the pitch and yaw components of gaze separately, we outperform recent work \cite{jeni2016person} in terms of gaze yaw ($3.13^\circ$ vs $3.51^\circ$), though perform worse in terms of gaze pitch ($6.92^\circ$ vs $4.27^\circ$).
This result is promising since \sysname{} operates in a dataset agnostic manner, while previous work \cite{jeni2016person} was trained on the Columbia dataset specifically.
Furthermore,
our second-order optimization strategy leads to faster convergence than first-order methods used in previous work \cite{wood20163d}, despite performing a similar amount of computation per iteration.
 
\subsection{Gaze redirection}

\begin{figure}[t]
    %
    %
    \includegraphics[width=1.0\linewidth]{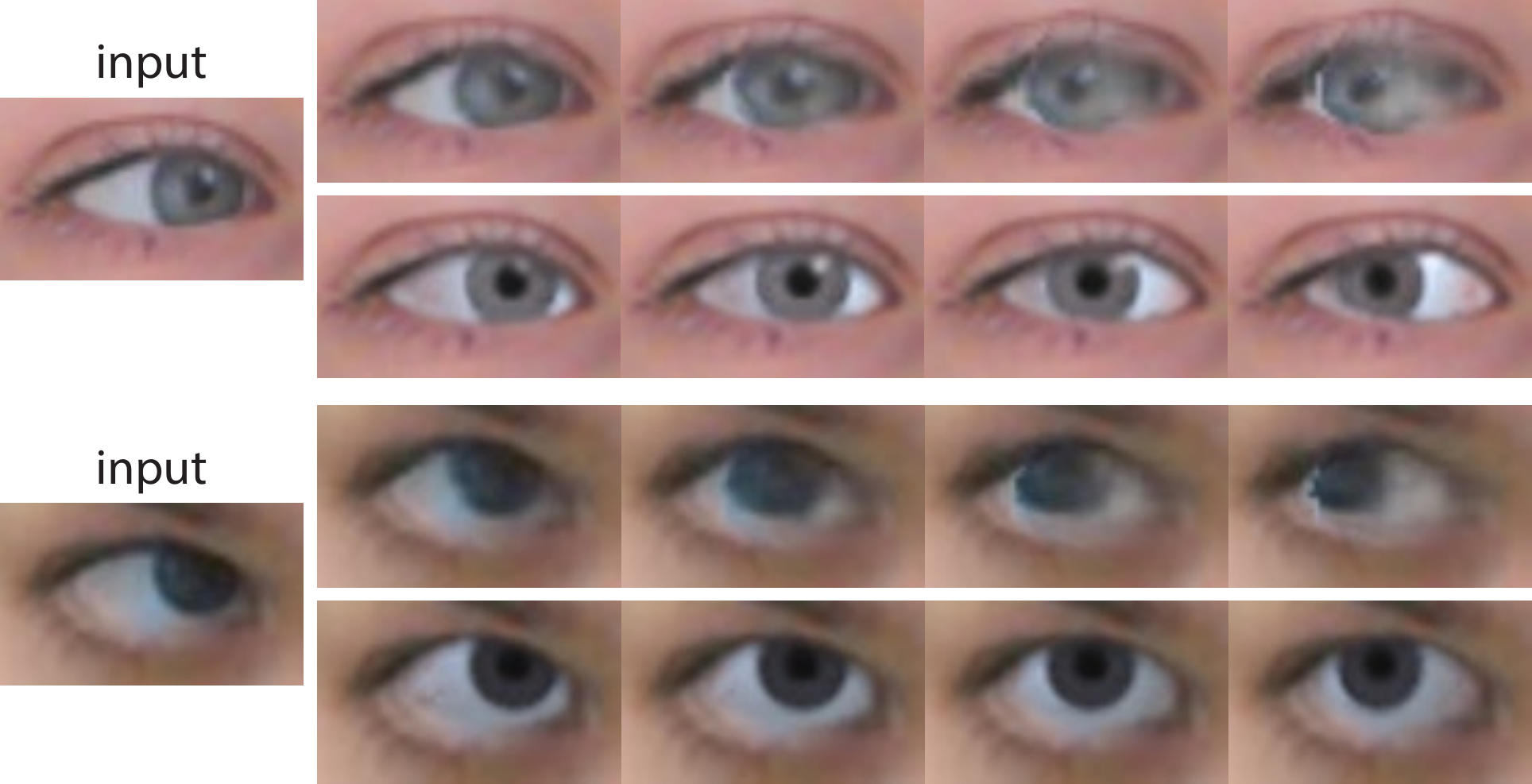}
    \caption{DeepWarp (top rows) \cite{ganin2016deepwarp} and \sysname{} (bottom rows) showing horizontal gaze redirection up to $45^\circ$. Our model based approach avoids the smudging artefacts encountered from large redirection angles with DeepWarp.}
    \label{fig:deepwarp}
\end{figure}

\begin{figure}
    \centering
    \includegraphics[width=\linewidth]{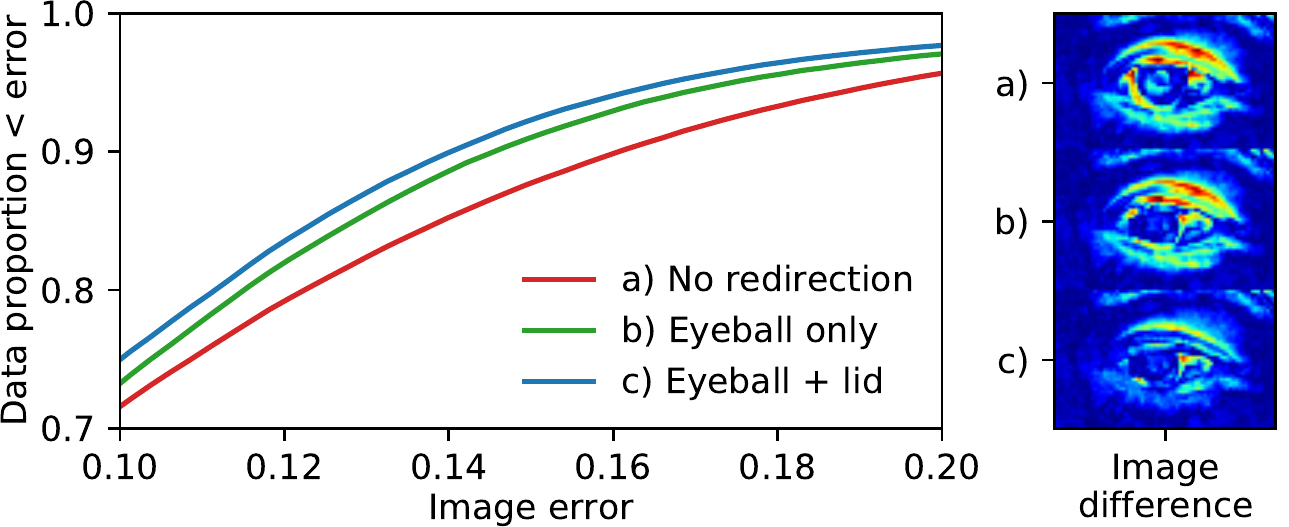} 
    \caption{Redirection error decreases as we enable more parts of our redirection pipeline. The $x$-axis represents image error, and the $y$-axis represents the proportion of data under that error.}
    \label{fig:exp-redirection}
\end{figure}

\begin{figure*}
    \makebox[0.134\linewidth]{\small \centering Input frame}
    \makebox[0.196\linewidth]{\small \centering $\pm 15^\circ$ pitch, $\pm 20^\circ$ yaw}
    \makebox[0.67\linewidth]{\small \centering Redirected eye gaze in YouTube videos} \par \smallskip
    \stackinset{l}{-1pt}{b}{-1pt}{\makebox[0pt][l]{\colorbox{white}{\small(c)}}}{%
    \stackinset{l}{-1pt}{b}{69pt}{\makebox[0pt][l]{\colorbox{white}{\small(b)}}}{%
    \stackinset{l}{-1pt}{b}{139pt}{\makebox[0pt][l]{\colorbox{white}{\small(a)}}}{%
    \includegraphics[width=\linewidth]{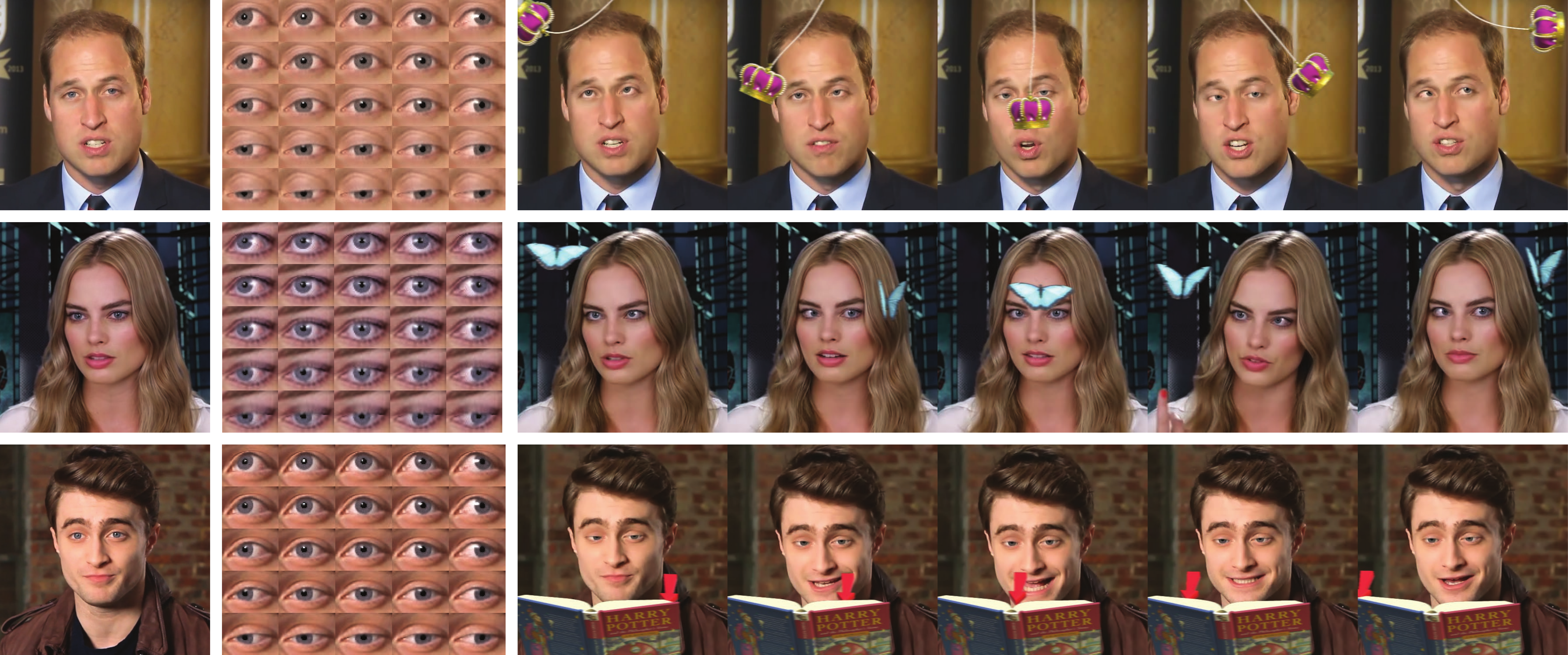}%
    }}}
    \caption{Example input frame, redirected eye gaze grid, and example output frames for three separate YouTube videos. (a,b): gaze has been redirected to new 3D gaze targets. (c): we have modified visual behaviour, making the video subject appear to read a book.}
    \label{fig:vidcaps}
\end{figure*}

We performed an experiment to evaluate our gaze redirection stages.
We prepared another subset of the Columbia gaze dataset \cite{Smith2013} with neutral head pose.
We aligned images of each participant using facial landmarks~\cite{Baltrusaitis2016}, and used the aligned images with different gaze as ground truth for ``redirected gaze''.
Following model fitting on the frontal gaze image, we produced three output images for each different gaze image:
a) with no gaze redirection,
b) with gaze redirection with the eyeballs only, and
c) with gaze redirection with eyeballs and eyelids.
We measured the per-pixel image difference between \sysname{} images and the ground truth redirected gaze images (see \autoref{fig:exp-redirection}).
The benefits of both eyeball redirection and eyelid redirection are clear.

\medskip
\textbf{Comparison to DeepWarp \cite{ganin2016deepwarp}} \quad
Previous work produces unsightly smudging artefacts when starting from non-central gaze, and redirecting gaze over large angles.
This is because their method fails to correctly hallucinate parts of the eyeball that were originally occluded.
As can be seen in \autoref{fig:deepwarp}, these issues do not arise with \sysname{} as we use a 3D model.
Furthermore, since DeepWarp can only apply an angular gaze offset to an input gaze direction, it cannot be used to produce results like those in \autoref{fig:vidcaps} where someone has been made to look at 3D gaze targets.
Please see our supplementary video for additional comparisons.

\subsection{Redirecting gaze in YouTube videos}

We demonstrate \sysname{} on videos with a variety of eye appearances, head pose, and illumination conditions by redirecting gaze in YouTube videos.
We downloaded videos from YouTube and resized them to a resolution of $640\!\times\!480$px.
New 3D gaze targets were specified through physics simulations and procedural programming using the Unity engine \cite{unity}.
\autoref{fig:vidcaps} shows some examples.
Please refer to our supplementary video for the full results.

\medskip
\textbf{Runtime} \quad \sysname{} runs on a commodity desktop machine ($3.3$Ghz CPU, NVidia GTX 1080).
Runtime is split between fitting and redirection.
We first process the entire video to recover $\bm\Phi^*$ for each frame.
This model fitting stage ran at 11.6fps, 12.5fps, and 12.1fps for the three YouTube videos in \autoref{fig:vidcaps}.
We then redirect gaze for each frame in the video.
Gaze redirection is less computationally demanding, and ran at 80fps for each video.


\section{Discussion}

In this work we described \sysname{}, a novel method for gaze redirection that uses model-fitting.
Unlike previous work, \sysname{} does not require person-specific training data, and can redirect eye gaze to new 3D targets explicitly.
We fit a parametric eye region model to images using analysis-by-synthesis, minimizing a reconstruction energy to recover shape, texture, pose, gaze, and illumination simultaneously.
Gaze redirection is then performed by warping eyelids, and compositing eyeballs onto the output in a photorealistic manner.

Limitations remain.
We do not explicitly model a full range of facial expressions such as blinking or squinting.
Furthermore, we do not handle occlusions or distortion effects from eyeglasses~\cite{Kubler2015}.
Our model does not include the eyelashes -- these are hard to model realistically, but can provide an important cue for downwards looking eye gaze.
We also do not consider cast shadows from hooded eyes or eyelashes.
Despite these limitations, we believe our work will enable a range of interesting and novel applications.

\subsection*{Acknowledgements}
This work was funded, in part, by the Cluster of Excellence on Multimodal Computing and Interaction at Saarland University, Germany.

{\small
\setlength{\bibsep}{0pt plus 0.3ex} 
\bibliographystyle{IEEEtranSN}		
\bibliography{refs}
}

\end{document}